\documentclass[11pt,a4paper]{article}
\usepackage[hyperref]{acl2020}
\usepackage{times}
\usepackage{latexsym}

\usepackage{url}

\aclfinalcopy 

\usepackage{multirow}
\usepackage{etoolbox}
\usepackage{booktabs}
\usepackage{tikz}
\usepackage{subfigure}
\usepackage{graphicx}
\usepackage{caption}
\usepackage{mwe}
\usepackage{epsfig}
\usepackage{tabularx}
\usepackage{multirow}
\usepackage[algoruled,boxed,noend]{algorithm2e}
\usepackage{mathtools}
\usepackage{multicol}

\usepackage{amsmath,amsfonts,bm}

\def\eqref#1{equation~\ref{#1}}

\def\1{\bm{1}}

\def\rvx{{\mathbf{x}}}
\def\rvy{{\mathbf{y}}}

\DeclareMathAlphabet{\mathsfit}{\encodingdefault}{\sfdefault}{m}{sl}
\SetMathAlphabet{\mathsfit}{bold}{\encodingdefault}{\sfdefault}{bx}{n}

\newcommand{\bA}{\mathbf{A}}

\newcommand{\bU}{\mathbf{U}}
\newcommand{\bM}{\mathbf{M}}
\newcommand{\bx}{\mathbf{x}}
\newcommand{\by}{\mathbf{y}}
\newcommand{\bh}{\mathbf{h}}

\newcommand{\bq}{\mathbf{q}}

\newcommand{\calL}{\mathcal{L}}
\newcommand{\calT}{\mathcal{T}}
\newcommand{\calX}{\mathcal{X}}

\newcommand{\PER}[1]{{\color{red}[\texttt{PER}~ #1]}}
\newcommand{\ORG}[1]{{\color{blue}[\texttt{ORG}~ #1]}}
\newcommand{\LOC}[1]{{\color{violet}[\texttt{LOC}~ #1]}}
\newcommand{\MISC}[1]{{\color{orange}[\texttt{MISC}~ #1]}}

\newcommand{\nop}[1]{}

\newcommand{\ie}{{\sl i.e.}}

\title{Learning to Contextually Aggregate \\ Multi-Source Supervision for Sequence Labeling}

\author{
Ouyu Lan\textsuperscript{\dag}\thanks{~~The first two authors contributed equally.}~~~
Xiao Huang\textsuperscript{\dag}$^*$~~
Bill Yuchen Lin\textsuperscript{\dag}~~
He Jiang\textsuperscript{\dag}~~
Liyuan Liu\textsuperscript{\ddag}~~
Xiang Ren\textsuperscript{\dag}\\
\texttt{\footnotesize\{olan,huan183,yuchen.lin,jian567,xiangren\}@usc.edu},~~\texttt{\footnotesize ll2@illinois.edu}
\\
\textsuperscript{\dag}{Computer Science Department, University of Southern California} \\ \textsuperscript{\ddag}Computer Science Department, University of Illinois at Urbana-Champaign
}

\date{}

\begin{document}
\maketitle

\begin{abstract}
    Sequence labeling is a fundamental framework for various natural language processing problems. Its performance is largely influenced by the annotation quality and quantity in supervised learning scenarios, and obtaining ground truth labels is often costly. In many cases, ground truth labels do not exist, but noisy annotations or annotations from different domains are accessible. In this paper, we propose a novel framework \textit{Consensus Network} (\textsc{ConNet}) that can be trained on annotations from multiple sources (e.g., crowd annotation, cross-domain data...). It learns individual representation for every source and dynamically aggregates source-specific knowledge by a context-aware attention module. Finally, it leads to a model reflecting the agreement (consensus) among multiple sources. We evaluate the proposed framework in two practical settings of multi-source learning: learning with crowd annotations and unsupervised cross-domain model adaptation. Extensive experimental results show that our model achieves significant improvements over existing methods in both settings. We also demonstrate that the method can apply to various tasks and cope with different encoders. \footnote{Our code can be found at \url{https://github.com/INK-USC/ConNet}~.} 
\end{abstract}
\section{Introduction}

\begin{figure*}[t]
    \centering
    \includegraphics[width=1.9\columnwidth]{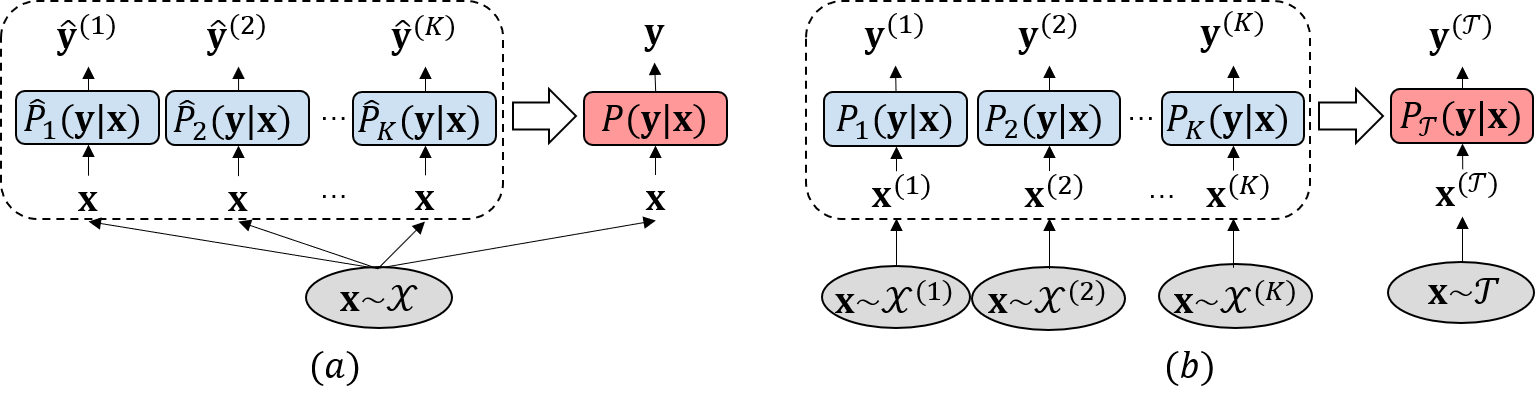}
    
    \caption{\textbf{Illustration of the task settings for the two applications in this work}:
    (a) learning consensus model from crowd annotations; (b) unsupervised cross-domain model adaptation.\vspace{-0.5cm}
    }
    \label{fig:prob}
\end{figure*}

Sequence labeling is a general approach encompassing various natural language processing (NLP) tasks including part-of-speech (POS) tagging~\citep{ratnaparkhi1996maximum}, word segmentation~\citep{low2005maximum}, and named entity recognition (NER)~\citep{nadeau2007survey}.
Typically, existing methods follow the supervised learning paradigm, and require high-quality annotations.
While gold standard annotation is expensive and time-consuming,
imperfect annotations are relatively easier to obtain from crowdsourcing (noisy labels) or other domains (out-of-domain).
Despite their low cost, such supervision usually can be obtained from different sources, and it has been shown that multi-source weak supervision has the potential to perform similar to gold annotations~\cite{ratner2016data}. 

Specifically, 
we are interested in two scenarios: 1) \textbf{learning with crowd annotations} and 2) \textbf{unsupervised cross-domain model adaptation}.
Both situations suffer from imperfect annotations, and benefit from multiple sources. 
Therefore, the key challenge here
is to aggregate multi-source imperfect annotations for learning a model without knowing the underlying ground truth label sequences in the target domain.


Our intuition mainly comes from the phenomenon that different sources of supervision have different strengths and are more proficient with distinct situations.
Therefore they may not keep consistent importance during aggregating supervisions, 
and aggregating multiple sources for a specific input should be a dynamic process that depends on the sentence context.
To better model this nature, we need to (1) explicitly model the unique traits of different sources when training and (2) find best suitable sources for generalizing the learned model on unseen sentences.

In this paper, we propose a novel framework, named \textit{Consensus Network} (\textsc{ConNet}), for sequence labeling with multi-source supervisions.
We represent the annotation patterns as different biases of annotators over a shared behavior pattern.
Both annotator-invariant patterns and annotator-specific biases are modeled in a decoupled way.
The first term comes through sharing part of low-level model parameters in a multi-task learning schema.
For learning the biases, we decouple them from the model as the transformations on top-level tagging model parameters, such that they can capture the unique strength of each annotator.
With such decoupled source representations,
we further learn an attention network for dynamically assigning the best sources for every unseen sentence through composing a transformation that represents the agreement among sources (consensus).
Extensive experimental results in two scenarios show that our model outperforms strong baseline methods, on various tasks and with different encoders. 
\textsc{ConNet} achieves state-of-the-art performance on real-world crowdsourcing datasets and improves significantly in unsupervised cross-domain adaptation tasks over existing works. 

\section{Related Work}
\label{sec:related}

There exists three threads of related work with this paper, which are sequence labeling, crowdsourcing and unsupervised domain adaptation.

\noindent \textbf{Neural Sequence Labeling.}
Traditional approaches for sequence labeling usually need significant efforts in feature engineering for graphical models like conditional random fields (CRFs)~\citep{Lafferty2001CRF}.
Recent research efforts in neural network models have shown that end-to-end learning like convolutional neural networks (CNNs)~\citep{ma2016end} or bidirectional long short-term memory (BLSTMs)~\citep{lample-etal-2016-neural} can largely eliminate human-crafted features.
BLSTM-CRF models have achieved promising performance~\citep{lample-etal-2016-neural} and are used as our base sequence tagging model in this paper.


\noindent \textbf{Crowd-sourced Annotation.}
Crowd-sourcing has been demonstrated to be an effective way of fulfilling the label consumption of neural models~\citep{Guan2017CoRR, Lin2019AlpacaTagAA}. 
It collects annotations with lower costs and a higher speed from non-expert contributors but suffers from some degradation in quality.
~\citet{Dawid1979EM} proposes the pioneering work to aggregate crowd annotations to estimate true labels, and~\citet{Snow2008EMNLP} shows its effectiveness with Amazon's Mechanical Turk system.
Later works~\citep{Dempster1977MLL,dredze2009sequence,raykar2010learning} focus on Expectation-Maximization (EM) algorithms to jointly learn the model and annotator behavior on classification.

Recent research shows the strength of multi-task framework in semi-supervised learning~\citep{lan2018semi,clark-etal-2018-semi}, cross-type learning~\citep{Wang2018BioNER}, and learning with entity triggers~\cite{TriggerNER2020}.
\citet{Nguyen2017ACL,Rodrigues2018AAAI,Simpson2020LowRS} regards crowd annotations as noisy gold labels and constructs crowd components to model annotator-specific bias which were discarded during the inference process. 
It is worth mentioning that, it has been found even for human curated annotations, there exists certain label noise that hinders the model performance~\cite{wang2019crossweigh}.
            
\noindent \textbf{Unsupervised Domain Adaptation.}
Unsupervised cross-domain adaptation aims to transfer knowledge learned from high-resource domains (source domains) to boost performance on low-resource domains (target domains) of interests such as social media messages~\cite{lin2017multi}.
Different from supervised adaptation~\citep{Lin2018NeuralAL}, we assume there is no labels at all for target corpora.
~\citet{saito2017asymmetric} and~\citet{ruder-plank-2018-strong} explored bootstrapping with multi-task tri-training approach, which requires unlabeled data from the target domain. 
The method is developed for one-to-one domain adaptation and does not model the differences among multiple source domains. 
~\citet{yang2015unsupervised} represents each domain with a vector of metadata domain attributes and uses domain vectors to train the model to deal with domain shifting, which is highly dependent on prior domain knowledge. ~\cite{ghifary2016deep} uses an auto-encoder method by jointly training a predictor for source labels, and a decoder to reproduce target input with a shared encoder. The decoder acts as a normalizer to force the model to learn shared knowledge between source and target domains. Adversarial penalty can be added to the loss function to make models learn domain-invariant feature only~\citep{fernando2015joint,long2014transfer,ming2015unsupervised}. However, it does not exploit domain-specific information. 
\section{Multi-source Supervised Learning}
\label{sec:fomrulation}
We formulate the multi-source sequence labeling problem as follows. 
Given $K$ sources of supervision, we regard each source 
as an imperfect annotator (non-expert human tagger or models trained in related domains).
For the $k$-th source data set {\small$S^{(k)}=\{(\rvx_i^{(k)}, \rvy_i^{(k)})\}_{i=1}^{m_k}$}, 
we denote its $i$-th sentence as $\rvx^{(k)}_{i}$ which is a sequence of tokens: {\small$\rvx^{(k)}_{i} = (x_{i, 1}^{(k)}, \cdots, x^{(k)}_{i, N})$}. 
The tag sequence of the sentence is 
marked as
{\small$\rvy^{(k)}_{i} = \{y_{i, j}^{{(k)}}\}$}.
We define the sentence set of each annotators as {\small$\mathcal{X}^{(k)} = \{\rvx_i^{(k)}\}_{i=1}^{m_k}$}, and the whole training domain as the union of all sentence sets: {\small$\mathcal{X} =  \bigcup_{k=1}^{(K)} \mathcal{X}^{(k)}$}.
The goal of the multi-source learning task is to use such imperfect annotations to train a model for predicting the tag sequence $\rvy$ for any sentence $\rvx$ in a target corpus $\mathcal{T}$.
Note that the target corpus $\mathcal{T}$ can either share the same distribution with $\mathcal{X}$ (Application I) or be significantly different (Application II).
In the following two subsections, we formulate two typical tasks in this problem as shown in Fig.~\ref{fig:prob}.


\smallskip
\noindent
\textbf{Application I: Learning with Crowd Annotations.}
When learning with crowd-sourced data, 
we regard each worker as an imperfect annotator ($S^{(k)}$), who may make mistakes or skip sentences in its annotations.
Note that in this setting, different annotators tag subsets of the \textit{same} given dataset ($\mathcal{X}$), and thus we assume there are no input distribution shifts among $\mathcal{X}^{(k)}$.
Also, we only test sentences in the same domain such that the distribution in target corpus $\mathcal{T}$ is the same as well.
That is, the marginal distribution of target corpus $P_{\mathcal{T}}(\rvx)$ is the same with that for each individual source dataset, \ie~$P_\mathcal{T}(\rvx) = P_k(\rvx)$. 
However, due to imperfectness of the annotations in each source, 
$P_k(\rvy | \rvx)$ 
is shifted from the underlying truth $P(\rvy|\rvx)$ (illustrated in the top-left part of  Fig.~\ref{fig:prob}).
The multi-source learning objective here is to learn a model $P_{\mathcal{T}}(\rvy|\rvx)$ for supporting inference on any new sentences in the same domain. 


\smallskip
\noindent
\textbf{Application II: Unsupervised Cross-Domain Model Adaptation.}
We assume 
there are available annotations in several source domains, but not in an unseen target domain. 
We assume
that the input distributions $P(\rvx)$ in different source domains $\mathcal{X}^{(k)}$ vary a lot,
and such annotations can hardly be adapted for training a target domain model. 
That is, the prediction distribution of each domain model ($P_k(\rvy|\rvx)$) is close to the underlying truth distribution ($P(\rvy|\rvx)$) only when $\rvx \in \mathcal{X}^{(k)}$.
For target corpus sentences $\rvx \in \mathcal{T}$, such a source model $P_k(\rvy|\rvx)$ again differs from underlying ground truth for the target domain $P_\mathcal{T}(\rvy|\rvx)$ and can be seen as an imperfect annotators.
Our objective in this setting is also to jointly model $P_\mathcal{T}(\rvy, \rvx)$ while noticing that there are significant domain shifts between $\mathcal{T}$ and any other $\mathcal{X}^{(k)}$.

\section{Consensus Network}
\label{sec:cn}

\label{sec:cn:overview}
\begin{figure*}[tbp]
\vspace{-0.3cm}
    \centering
    \includegraphics[width=1.7\columnwidth]{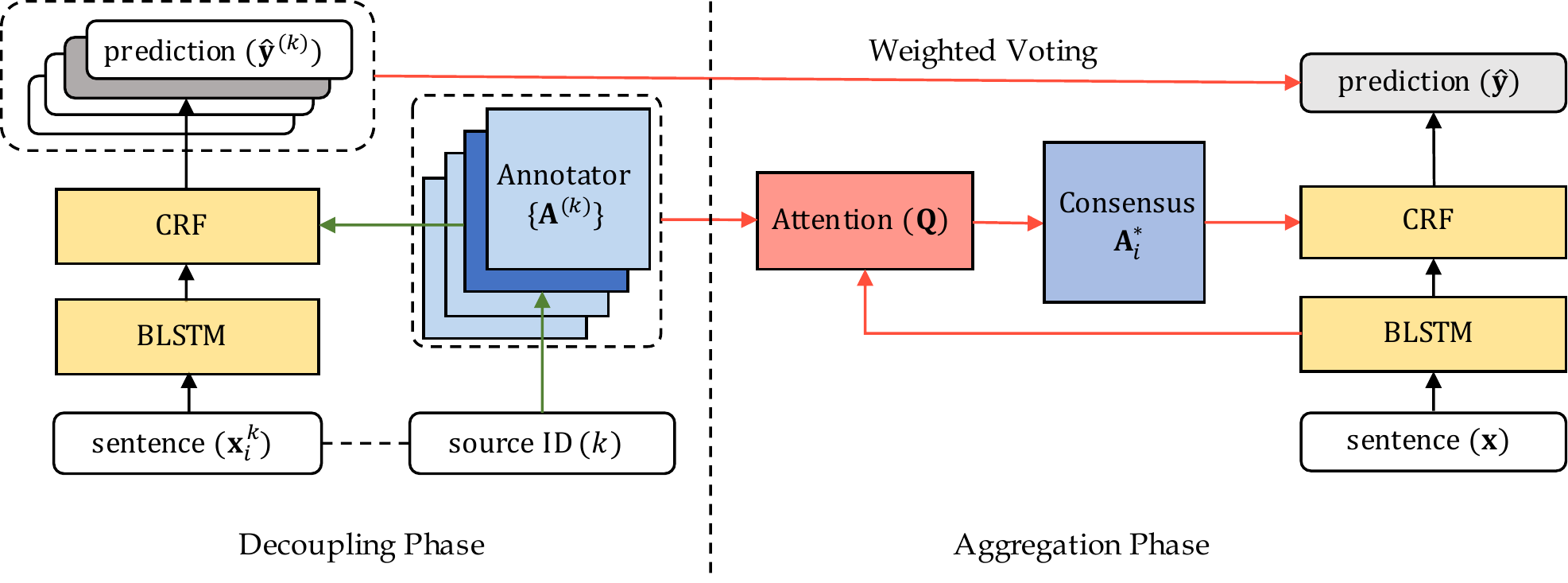}
    \vspace{-0.3cm}
    \caption{\small \textbf{Overview of the \textsc{ConNet} framework.} 
    The decoupling phase constructs the shared model (yellow) and source-specific matrices (blue). The aggregation phase dynamically combines crowd components into a consensus representation (blue) by a context-aware attention module (red) for each sentence $x$. 
    \vspace{-15pt}
    }
    \label{fig:cn}
\end{figure*}

In this section, we present our two-phase framework \textsc{ConNet} for multi-source sequence labeling. As shown in Figure~\ref{fig:cn}, our proposed framework first uses a multi-task learning schema with a special objective to decouple annotator representations as different parameters of a transformation around CRF layers.
This \textbf{decoupling phase} (Section~\ref{sec:cn:ep}) is for decoupling the model parameters into a set of annotator-invariant model parameters and a set of annotator-specific representations.
Secondly, the dynamic \textbf{aggregation phase} (Section~\ref{sec:cn:ap}) learns to contextually utilize the annotator representations with a lightweight attention mechanism to find the best suitable transformation for each sentence, so that the model can achieve a context-aware consensus among all sources. The inference process is described in Section~\ref{sec:cn:inf}.

\subsection{The Base Model: BLSTM-CRF}
\label{sec:base}
Many recent sequence labeling frameworks~\citep{Ma2016ACL,misawa-etal-2017-character} share a very basic structure: a bidirectional LSTM network followed by a CRF tagging layer (i.e. BLSTM-CRF).
The BLSTM encodes an input sequence $\bx = \{x_1, x_2, \ldots, x_n\}$ into a sequence of hidden state vectors $\mathbf{h}_{1:n}$.
The CRF takes as input the hidden state vectors and computes an emission score matrix $\bU \in\mathbb{R}^{n\times L}$ where $L$ is the size of tag set. It also maintains a trainable transition matrix $\bM \in\mathbb{R}^{L\times L}$. We can consider $\bU_{i,j}$ is the score of labeling the tag with id $j\in \{1,2,...,L\}$ for $i^{th}$ word in the input sequence $\bx$, and $\bM_{i,j}$ means the transition score from $i^{th}$ tag to $j^{th}$.

The CRF further computes the score $s$ for a predicted tag sequence $\by=\{y_1, y_2, ..., y_k\}$ as 
\begin{equation}
    s({\bx},{\by}) = \sum_{t=1}^T (\bU_{t, {y}_{t}} + \bM_{{y}_{t-1},{y}_t}),
\label{eq:crf}
\end{equation}
and then tag sequence $\rvy$ follows the conditional distribution
\vspace{-5pt}
\begin{equation}
\vspace{-5pt}
    P({\by} | \mathbf{x})=\frac{\exp{s(\mathbf{x}, {\by})}}{\sum_{{{\by’}} \in {Y}_{\mathbf{x}}} \exp{s(\mathbf{x}, {{\by‘}})}}.
    \label{softmax}
\end{equation}

\subsection{The Decoupling Phase: Learning annotator representations}
\label{sec:cn:ep}
For decoupling annotator-specific biases in annotations, we represent them as a transformation on emission scores and transition scores respectively.
Specifically, we learn a matrix $\bA^{(k)} \in \mathbb{R}^{L\times L}$ for each imperfect annotator $k$ and apply this matrix as transformation on $\bU$ and $\bM$ as follows:
\begin{equation}
 s^{(k)}({\bx},{\by}) = \sum_{t=1}^T \left((\bU \bA^{(k)})_{t, {y}_t} + (\bM \bA^{(k)})_{{y}_{t-1},{y}_t}\right).
\label{eq:df:crf}
\end{equation}
From this transformation, we can see that the original score function $s$ in Eq.~\ref{eq:crf} becomes an source-specific computation. 
The original emission and transformation score matrix $\bU$ and $\bM$ are still shared by all the annotators, while they both are transformed by the matrix $\bA^{(k)}$ for $k$-th annotator.
While training the model parameters in this phase,
we follow a multi-task learning schema. 
That is, we share the model parameters for BLSTM and CRF (including $\mathbf{W}$, $\mathbf{b}$, $\bM$), while updating $\bA^{(k)}$ only by examples in $S_k = \{\mathcal{X}^{(k)}, \mathcal{Y}^{(k)}\}$.

The learning objective is to minimize the negative log-likelihood of all source annotations:
\begin{align}
\calL =& - \log \sum_{k=1}^{K} \sum_{i=1}^{|\mathcal{X}^{(k)}|} P(\rvy_i^{(k)}| \rvx_i^{(k)})~~,
\label{eq:basic:loss} 
\end{align}
\vspace{-15pt}
\begin{align}
P(\rvy_i^{(k)}| \rvx_i^{(k)}) = \frac{\exp{s^{(k)}({\rvx_i^{(k)}},{\rvy_i^{(k)}})}}{\sum_{\rvy'} \exp{s^{(k)}({\bx},{\rvy'})}}.
\label{eq:basic:p}
\end{align}
The assumption on the annotation representation $\bA^{(k)}$ is that it can model the pattern of annotation bias.
Each annotator can be seen as a noisy version of the shared model. For the $k$-th annotator, $\bA^{(k)}$ models noise from labeling the current word and transferring from the previous label. Specifically, each entry $\bA^{(k)}_{i, j}$ captures the probability of mistakenly labeling $i$-th tag to $j$-th tag. In other words, the base sequence labeling model in Sec.~\ref{sec:base} learns the basic consensus knowledge while annotator-specific components add their understanding to predictions.

\subsection{The Aggregation Phase: Dynamically Reaching Consensus}
\label{sec:cn:ap}

In the second phase, our proposed network learns a context-aware attention module for a consensus representation supervised by combined predictions on the target data.
For each sentence in target data $\mathcal{T}$, these predictions are combined by weighted voting.
The weight of each source is its normalized $F_1$ score on the training set.
Through weighted voting on such augmented labels over all source sentences $\mathcal{X}$, 
we can find a good approximation of underlying truth labels. 

For better generalization and higher speed, an attention module is trained to estimate the relevance of each source to the target under the supervision of generated labels. 
Specifically, we compute the sentence embedding by concatenating the last hidden states of the forward LSTM and the backward LSTM, \ie~ $\bh^{(i)} = [\overrightarrow{\bh}_T^{(i)}; \overleftarrow{\bh}_0^{(i)}]$. The attention module inputs the sentence embedding and outputs a normalized weight for each source:
\begin{align}
    \mathbf{q}_i= \text{softmax}(\mathbf{Q} \mathbf{h}^{(i)}),\text{~~where~~}\mathbf{Q}\in\mathbb{R}^{K \times 2d}.
\label{eq:attvec}
\end{align}
where $d$ is the size of each hidden state $\mathbf{h}^{(i)}$.
Source-specific matrices $\{\bA^{(k)}\}_{k=1}^K$ are then aggregated into a consensus representation $\mathbf{A}_i^*$ for sentence $\bx_i \in \calX$ by
\begin{align}
    \mathbf{A}_i^* = \sum_{k=1}^K {q}_{i,k} \bA^{(k)}.
\vspace{-3mm}
\end{align}
In this way, the consensus representation contains more information about sources which are more related to the current sentence. 
It also alleviates the contradiction problem among sources, because it could consider multiple sources of different emphasis.
Since only an attention model with weight matrix $\mathbf{Q}$ is required to be trained, the amount of computation is relatively small.
We assume the base model and annotator representations are well-trained in the previous phase. The main objective in this phase is to learn how to select most suitable annotators for the current sentence.

\subsection{Parameter Learning and Inference}
\label{sec:cn:inf}
\textsc{ConNet} learns parameters through two phases described above. In the decoupling phase, each instance from source $S_k$ is used for training the base sequence labeling model and its representation $\bA^{(k)}$. 
In the aggregation phase, we use aggregated predictions from the first phase to learn a lightweight attention module.
For each instance in the target corpus $\bx_i \in \calT$, we calculate its embedding $\bh_i$ from BLSTM hidden states. 
With these sentence embeddings, the context-aware attention module assigns weight $\mathbf{q}_i$ to each source and dynamically aggregates source-specific representations $\{\bA^{(k)}\}$ for inferring $\hat{\by}_i$. In the inference process, only the consolidated consensus matrix $\bA^{*}_i$ is applied to the base sequence learning model. In this way, more specialist knowledge helps to deal with more complex instances.

\subsection{Model Application}
\label{sec:cn:app}
The proposed model can be applied to two practical multi-sourcing settings: learning with crowd annotations and unsupervised cross-domain model adaptation.  In the crowd annotation learning setting, the training data of the same domain is annotated by multiple noisy annotators, and each annotator is treated as a source. In the decoupling phase, the model is trained on noisy annotations, and in the aggregation phase, it is trained with combined predictions on the training set. In the cross-domain setting, the model has access to unlabeled training data of the target domain and clean labeled data of multiple source domains. Each domain is treated as a source. In the decoupling phase, the model is trained on source domains, and in the aggregation phase, the model is trained on combined predictions on the training data of the target domain.
Our framework can also extend to new tasks other than sequence labeling and cope with different encoders. We will demonstrate this ability in experiments.

Our method is also incorporated as a feature for controlling the quality of crowd-annotation  in annotation frameworks such as AlpacaTag~\cite{Lin2019AlpacaTagAA} and LEAN-LIFE~\cite{LEANLIFE}.
\section{Experiments}





We evaluate \textsc{ConNet} in the two aforementioned settings of multi-source learning: learning with crowd annotations and unsupervised cross-domain model adaptation.
Additionally, to demonstrate the generalization of our framework, we also test our method on sequence labeling with transformer encoder in Appendix \ref{sec:app:tf} and text classification with MLP encoder in Section \ref{sec:exp:domain}.

\begin{table*}
\centering
\scalebox{0.7}{
\begin{tabular}{lcccccc}
    \toprule  
    \multirow{2}{*}{\textbf{Methods}}&
    \multicolumn{3}{c}{ \textbf{AMTC}}&\multicolumn{3}{c}{ \textbf{AMT}}\\
    \cmidrule(lr){2-4} \cmidrule(lr){5-7}
     & Precision(\%) & Recall(\%) & F1-score(\%) & Precision(\%) & Recall(\%) & F1-score(\%)  \\
    \midrule  
    {CONCAT-SLM}  &  \textbf{85.95}($\pm$1.00)  &  57.96($\pm$0.26)  &  69.23($\pm$0.13) & \textbf{91.12}($\pm$0.57) & 55.41($\pm$2.66) & 68.89($\pm$1.92) \\
    {MVT-SLM}  &  84.78($\pm$0.66)  &  62.50($\pm$1.36)  &  71.94($\pm$0.66) & 86.96($\pm$1.22) & 58.07($\pm$0.11) & 69.64($\pm$0.31) \\
    {MVS-SLM} &  84.76($\pm$0.50)  &  61.95($\pm$0.32)  &  71.57($\pm$0.04) & 86.95($\pm$1.12) & 56.23($\pm$0.01) & 68.30($\pm$0.33) \\
    {DS-SLM}~\citep{Nguyen2017ACL}  &  72.30$^*$  &  61.17$^*$  &  66.27$^*$ & - & - & - \\
    {HMM-SLM}~\citep{Nguyen2017ACL} &  76.19$^*$  &  66.24$^*$  &  70.87$^*$ & - & - & - \\
    {MTL-MVT}~\citep{Wang2018BioNER}  & 81.81($\pm$2.34) & 62.51($\pm$0.28) & 70.87($\pm$1.06) &  88.88($\pm$0.25)  & 65.04($\pm$0.80)   & 75.10($\pm$0.44)   \\
    {MTL-BEA}~\citep{rahimi2019massively} & 85.72($\pm$0.66) & 58.28($\pm$0.43) & 69.39($\pm$0.52) & 77.56($\pm$2.23) & 67.23($\pm$0.72) & 72.01($\pm$0.85) \\
    \midrule
    {CRF-MA}~\citep{Rodrigues2014ML} & -  &  -  &  - &  49.40$^*$  &  \textbf{85.60}$^*$  &  62.60$^*$  \\
    {Crowd-Add}~\citep{Nguyen2017ACL}  &  85.81($\pm$1.53)  &  62.15($\pm$0.18)  &  72.09($\pm$0.42) & 89.74($\pm$0.10) & 64.50($\pm$1.48) & 75.03($\pm$1.02) \\
    {Crowd-Cat}~\citep{Nguyen2017ACL}  & 85.02($\pm$0.98) & 62.73($\pm$1.10) & 72.19($\pm$0.37) &  89.72($\pm$0.47) & 63.55($\pm$1.20) & 74.39($\pm$0.98) \\
    {CL-MW}~\citep{Rodrigues2018AAAI}  & - & - & - &  66.00$^*$  &  59.30$^*$  &  62.40$^*$  \\
    \midrule
    \textsc{ConNet} (Ours) &  84.11($\pm$0.71)  &  \textbf{68.61}($\pm$0.03)  &  \textbf{75.57}($\pm$0.27) & 88.77($\pm$0.25) & 72.79($\pm$0.04) & \textbf{79.99}($\pm$0.08) \\
    \midrule
    {Gold} (Upper Bound) &  89.48($\pm$0.32)  &  89.55($\pm$0.06)  &  89.51($\pm$0.21) & 92.12($\pm$0.31) & 91.73($\pm$0.09) & 91.92($\pm$0.21) \\
    \bottomrule 
\end{tabular}
}
\caption{\small{\textbf{Performance on real-world crowd-sourced NER datasets.} 
The best score in each column excepting \texttt{Gold} is marked \textbf{bold}. * indicates number reported by the paper.}
\vspace{-10pt}
}
\label{tab:res:real}
\end{table*}

\subsection{Datasets}
\label{sec:exp:data}



\textbf{Crowd-Annotation Datasets.}
We use crowd-annotation datasets based on the 2003 CoNLL shared NER task ~\citep{tjong-kim-sang-de-meulder-2003-introduction}. The real-world datasets, denoted as AMT, are collected by~\citet{Rodrigues2014ML} using Amazon's Mechanical Turk where F1 scores of annotators against the ground truth vary from 17.60\% to 89.11\%. Since there is no development set in AMT, we also follow~\citet{Nguyen2017ACL} to use the AMT training set and CoNLL 2003 development and test sets, denoted as AMTC. Overlapping sentences are removed in the training set, which is ignored in that work. 
Additionally, we construct two sets of simulated datasets to investigate the quality and quantity of annotators. To simulate the behavior of a non-expert annotator, a CRF model is trained on a small subset of training data and generates predictions on the whole set. Because of the limited size of training data, each model would have a bias to certain patterns.

\smallskip
\noindent
\textbf{Cross-Domain Datasets.}
In this setting, we investigate three NLP tasks: POS tagging, NER and text classification.
For POS tagging task, we use the GUM portion~\citep{zeldes2017gum} of Universal Dependencies (UD) v2.3 corpus with $17$ tags and $7$ domains: academic, bio, fiction, news, voyage, wiki, and interview. 
For NER task, we select the English portion of the OntoNotes v5 corpus~\citep{hovy2006ontonotes}. The corpus is annotated with $9$ named entities with data from $6$ domains: broadcast conversation (bc), broadcast news (bn), magazine (mz), newswire (nw), pivot text (pt), telephone conversation (tc), and web (web). 
Multi-Domain Sentiment Dataset (MDS) v2.0~\citep{blitzer2007biographies} is used for text classification, which is built on Amazon reviews from $4$ domains: books, dvd, electronics, and kitchen. Since the dataset only contains word frequencies for each review without raw texts, we follow the setting in~\citet{chen-cardie-2018-multinomial} considering 5,000 most frequent words and use the raw counts as the feature vector for each review.


\subsection{Experiment Setup}
\label{sec:exp:setup}

For sequence labeling tasks, we follow \citet{liu2018efficient} to build the BLSTM-CRF architecture as the base model. The dimension of character-level, word-level embeddings and LSTM hidden layer are set as $30$, $100$ and $150$ respectively.
For text classification, each review is represented as a $5000$-d vector. We use an MLP with a hidden size of $100$ for encoding features and a linear classification layer for predicting labels.
The dropout with a probability of $0.5$ is applied to the non-recurrent connections for regularization.
The network parameters are updated by stochastic gradient descent (SGD).
The learning rate is initialized as $0.015$ and decayed by $5\%$ for each epoch.
The training process stops early if no improvements in $15$ continuous epochs and selects the best model on the development set.
For the dataset without a development set, we report the performance on the $50$-th epoch.
For each experiment, we report the average performance and standard variance of $3$ runs with different random initialization.

\subsection{Compared Methods}
\label{sec:exp:comp}

\begin{figure*}
\vspace{-0.3cm}
    \centering
    \includegraphics[width=1\linewidth]{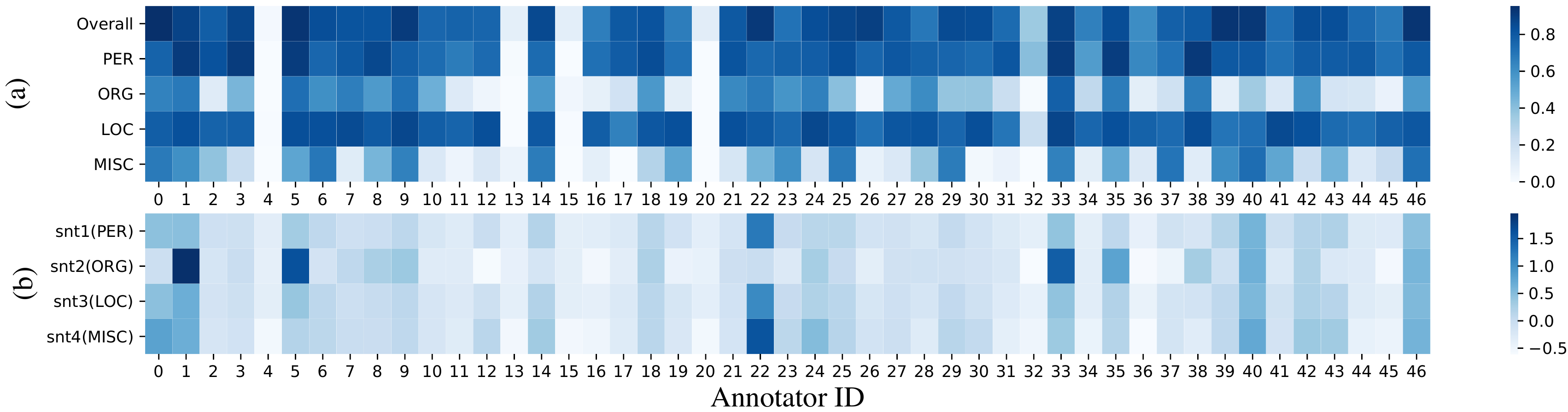}
    \vspace{-0.6cm}
    \caption{\small \textbf{Visualizations of (a) the expertise of annotators; (b) attention weights for sample sentences.}
    More cases and details are described in Appendix~\ref{sec:app:case}. 
    \vspace{-0.3cm}
    }
    \label{fig:exp:crowd:caseheat}
\end{figure*}

We compare our models with multiple baselines, which can be categorized in two groups: wrapper methods and joint models. To demonstrate the theoretical upper bound of performance, we also train the base model using ground-truth annotations in the target domain (\texttt{Gold}).

A wrapper method consists of a label aggregator and a deep learning model. These two components could be combined in two ways: (1) aggregating labels on crowd-sourced training set then feeding the generated labels to a Sequence Labeling Model (\texttt{SLM})~\citep{Liu2017EMNLP}; (2) feeding multi-source data to a Multi-Task Learning (\texttt{MTL})~\citep{Wang2018BioNER} model then aggregating multiple predicted labels. 
We investigate multiple label aggregation strategies. \texttt{CONCAT} considers all crowd annotations as gold labels. \texttt{MVT} does majority voting on the token level, \ie, the majority of labels $\{\by_{i,j}^k\}$ is selected as the gold label for each token $\bx_{i,j}$. \texttt{MVS} is conducted on the sequence level, addressing the problem of violating Begin/In/Out (BIO) rules. \texttt{DS}~\citep{Dawid1979EM}, \texttt{HMM}~\citep{Nguyen2017ACL} and \texttt{BEA}~\citep{rahimi2019massively} induce consensus labels with probability models.

In contrast with wrapper methods, joint models incorporate multi-source data within the structure of sequential taggers and jointly model all individual annotators. \texttt{CRF-MA} models CRFs with Multiple Annotators by EM algorithm~\citep{Rodrigues2014ML}.
\citet{Nguyen2017ACL} augments the LSTM architecture with crowd vectors.
These crowd components are element-wise added to tags scores (\texttt{Crowd-Add}) or concatenated to the output of hidden layer (\texttt{Crowd-Cat}).
These two methods are the most similar to our decoupling phase. We implemented them and got better results than reported.
\texttt{CL-MW} applies a crowd layer to a CNN-based deep learning framework \citep{Rodrigues2018AAAI}. 
\texttt{Tri-Training} uses bootstrapping with multi-task Tri-Training approach for unsupervised one-to-one domain adaptation~\citep{saito2017asymmetric,ruder-plank-2018-strong}.

\subsection{Learning with Crowd Annotations}

\noindent\textbf{Performance on real-world datasets.} Tab.~\ref{tab:res:real} shows the performance of aforementioned methods and our \textsc{ConNet} on two real-world datasets, \ie~AMT and AMTC\footnote{\scriptsize We tried our best to re-implement the baseline methods for all datasets, and left the results blank when the re-implementation is not showing consistent results as in the original papers.}.
We can see that \textsc{ConNet} outperforms all other methods on both datasets significantly on $F1$ score, which shows the effectiveness of dealing with noisy annotations for higher-quality labels.
Although \texttt{CONCAT-SLM} achieves the highest precision, it suffers from low recall.
Most existing methods have the high-precision but low-recall problem.
One possible reason
is that they try to find the latent ground truth and throw away illuminating annotator-specific information. 
So only simple mentions can be classified with great certainty while difficult mentions fail to be identified without sufficient knowledge.
In comparison, \textsc{ConNet} pools information from all annotations and focus on matching knowledge to make predictions. 
It makes the model be able to identify more mentions and get a higher recall.


\begin{figure}[tbp]
\vspace{-0.1cm}
\centering
\includegraphics[width=0.8\linewidth]{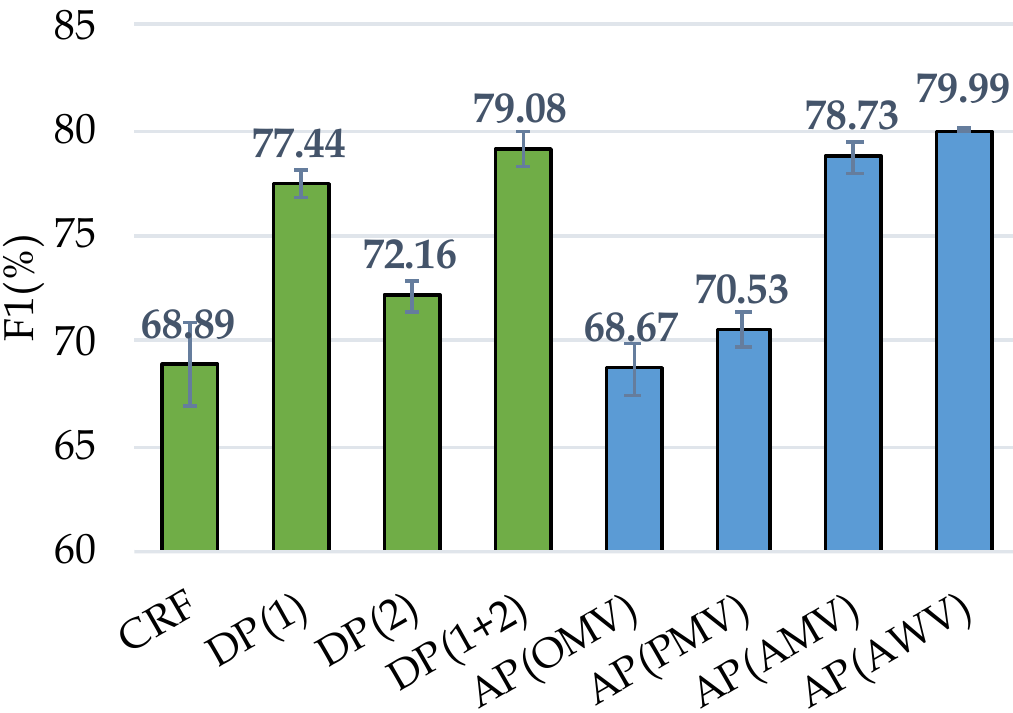}
\vspace{-0.2cm}
\caption{\small \textbf{Performance of \textsc{ConNet} variants} of decoupling phase (DP) and aggregation phase (AP).}
\label{fig:exp:crowd:ab}
\vspace{-0.2cm}
\end{figure}

\noindent\textbf{Case study.} It is enlightening to analyze whether the model decides the importance of annotators given a sentence. Fig.~\ref{fig:exp:crowd:caseheat} visualizes test F1 score of all annotators, and attention weights $\bq_i$ in Eq.~\ref{eq:attvec} for $4$ sampled sentences containing different entity types. Obviously, the $2$nd sample sentence with \texttt{ORG} has higher attention weights on $1$st, $5$th and $33$rd annotator who are best at labeling \texttt{ORG}.
More details and cases are shown in Appendix~\ref{sec:app:case}. 

\begin{figure}
\vspace{-0.15cm}
\hspace{-0.3cm}
\includegraphics[width=1.05\linewidth]{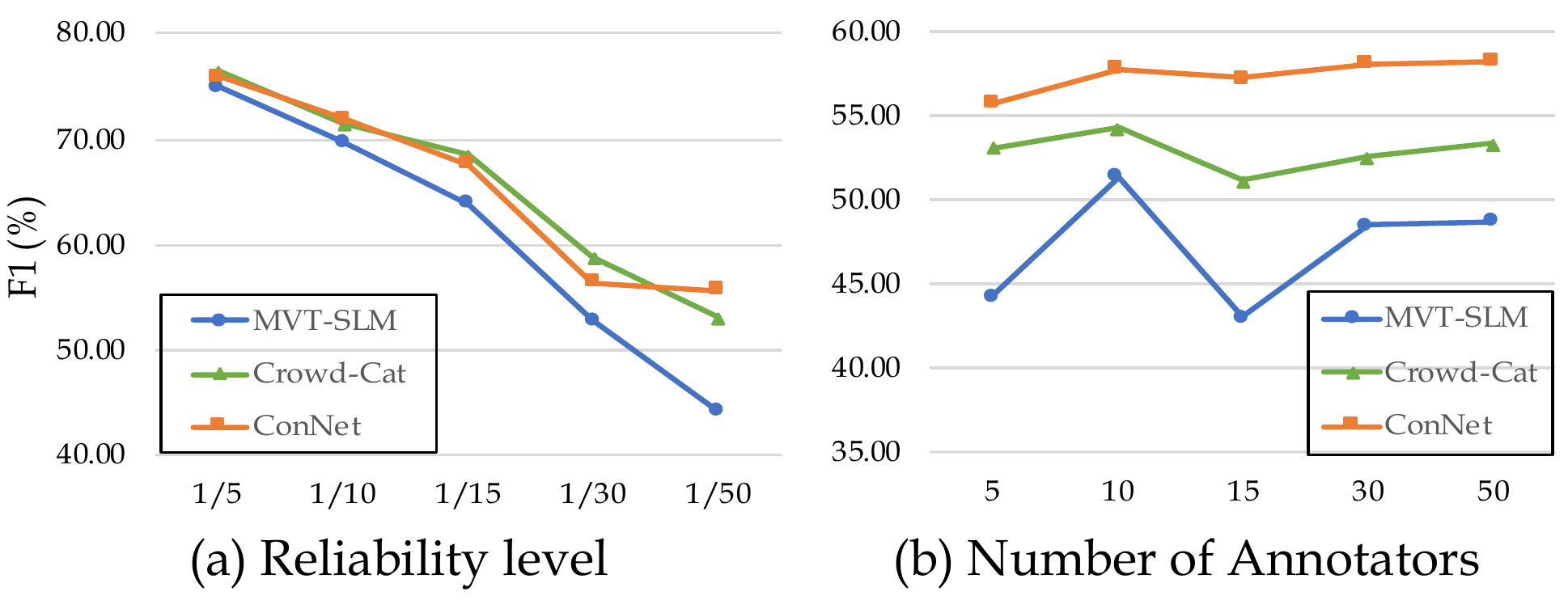}
\vspace{-0.6cm}
\caption{\small \textbf{Performance on simulated crowd-sourced NER data} with (a) $5$ annotators with different reliability levels; (b) different numbers of annotators with reliability $r=1/50$. }
\label{fig:exp:crowd:simu}
\vspace{-0.4cm}
\end{figure}

\begin{table*}[t]
\centering
\scalebox{0.7}{
\begin{tabularx}{1.3\textwidth}{l|c c c c c c c X|r}
  \toprule
  Task \& Corpus & \multicolumn{9}{c}{\bf Multi-Domain POS Tagging: Universal Dependencies v2.3 - GUM} \\
  \midrule
  Target Domain & academic & bio~~ & ~fiction~ & ~~news~ & voyage~ & wiki & interview & & AVG $Acc$(\%) \\
  \midrule
  CONCAT & 92.68 & 92.12 & 93.05 & 90.79 & 92.38 & 92.32 & 91.44 & & 92.11($\pm$0.07) \\
  MTL-MVT~\citep{Wang2018BioNER} & 92.42 & 90.59 & 91.16 & 89.69 & 90.75 & 90.29 & 90.21 & & 90.73($\pm$0.29) \\
  MTL-BEA~\citep{rahimi2019massively} & 92.87 & 91.88 & 91.90 & 91.03 & 91.67 & 91.31 & 91.29 & & 91.71($\pm$0.06) \\
  \midrule
  Crowd-Add~\citep{Nguyen2017ACL} & 92.58 & 91.91 & 91.50 & 90.73 & 91.74 & 90.47 & 90.61 & & 91.36($\pm$0.14) \\
  Crowd-Cat~\citep{Nguyen2017ACL} & 92.71 & 91.71 & 92.48 & 91.15 & 92.35 & 91.97 & 91.22 & & 91.94($\pm$0.08) \\
  Tri-Training~\citep{ruder-plank-2018-strong} & 92.84 & 92.15 & 92.51 & \underline{91.40} & 92.35 & 91.29 & 91.00 & & 91.93($\pm$0.01) \\
  \midrule
  \textsc{ConNet} & \underline{92.97} & \underline{92.25} & \underline{93.15} & 91.06 & \underline{92.52} & \textbf{92.74} & \underline{91.66} & & \underline{92.33}($\pm$0.17) \\
  \midrule
  Gold (Upper Bound) & 92.64 & 93.10 & 93.15 & 91.33 & 93.09 & 94.67 & 92.20 & & 92.88($\pm$0.14) \\
  \bottomrule
\end{tabularx}
}
\scalebox{0.7}{
\begin{tabularx}{1.3\textwidth}{l|X c X c X c X c X c X c X|l}
  \toprule
  Task \& Corpus & \multicolumn{14}{c}{\bf Multi-Domain NER: OntoNotes v5.0 - English} \\
  \midrule
  Target Domain & & nw & & wb & & bn & & tc & & bc & & mz & & AVG $F_1$(\%) \\
  \midrule
  CONCAT & & 68.23 & & 32.96 & & 77.25 & & \underline{53.66} & & 72.74 & & 62.61 & & 61.24($\pm$0.92) \\
  MTL-MVT~\citep{Wang2018BioNER} & & 65.74 & & 33.25 & & 76.80 & & 53.16 & & 69.77 & & 63.91 & & 60.44($\pm$0.45) \\
  MTL-BEA~\citep{rahimi2019massively} & & 58.33 & & 32.62 & & 72.47 & & 47.83 & & 48.99 & & 52.68 & & 52.15($\pm$0.58) \\
  \midrule
  Crowd-Add~\citep{Nguyen2017ACL} & & 45.76 & & 32.51 & & 50.01 & & 26.47 & & 52.94 & & 28.12 & & 39.30($\pm$4.44) \\
  Crowd-Cat~\citep{Nguyen2017ACL} & & 68.95 & & 32.61 & & 78.07 & & 53.41 & & \textbf{74.22} & & 65.55 & & 62.14($\pm$0.89) \\
  Tri-Training~\citep{ruder-plank-2018-strong} & & 69.68 & & 33.41 & & 79.62 & & 47.91 & & 70.85 & & 68.53 & & 61.67($\pm$0.31) \\
  \midrule
  \textsc{ConNet} & & \textbf{71.31} & & \textbf{34.06} & & \underline{79.66} & & 52.72 & & 71.47 & & \textbf{70.71} & & \textbf{63.32}($\pm$0.81) \\
  \midrule
  Gold (Upper Bound) & & 84.70 & & 46.98 & & 83.77 & & 52.57 & & 73.05 & & 70.58 & & 68.61($\pm$0.64) \\
  \bottomrule
\end{tabularx}
}
\scalebox{0.7}{
\begin{tabularx}{1.3\textwidth}{l|X c X c X c X c X|l}
  \toprule
  Task \& Corpus & \multicolumn{10}{c}{\bf Multi-Domain Text Classification: MDS} \\
  \midrule
  Target Domain & & books & & dvd & & electronics & & kitchen & & AVG $Acc$(\%) \\
  \midrule
  CONCAT & & 75.68 & & 77.02 & & 81.87 & & 83.07 & & 79.41($\pm$0.02) \\
  MTL-MVT~\citep{Wang2018BioNER} & & 74.92 & & 74.43 & & 79.33 & & 81.47 & & 77.54($\pm$0.06) \\
  MTL-BEA~\citep{rahimi2019massively} & & 74.88 & & 74.60 & & 79.73 & & 82.82 & & 78.01($\pm$0.28) \\
  \midrule
  Crowd-Add~\citep{Nguyen2017ACL} & & 75.72 & & 77.35 & & 81.25 & & 82.90 & & 79.30($\pm$9.21) \\
  Crowd-Cat~\citep{Nguyen2017ACL} & & 76.45 & & 77.37 & & 81.22 & & 83.12 & & 79.54($\pm$0.25) \\
  Tri-Training~\citep{ruder-plank-2018-strong} & & 77.58 & & 78.45 & & 81.95 & & 83.17 & & 80.29($\pm$0.02) \\
  \midrule
  \textsc{ConNet} & & \textbf{78.75} & & \textbf{81.06} & & \textbf{84.12} & & \underline{83.45} & & \textbf{81.85}($\pm$0.04) \\
  \midrule
  Gold (Upper Bound) & & 78.78 & & 82.11 & & 86.21 & & 85.76 & & 83.22($\pm$0.19) \\
  \bottomrule
\end{tabularx}
}
\caption[]{\textbf{Performance on cross-domain data
} The best score (except the Gold) in each column that is significantly ($p<0.05$) better than the second best is marked \textbf{bold}, while those are better but not significantly are \underline{underlined}.
}
\label{tab:app:domain}
\end{table*}
\noindent\textbf{Ablation study.} We also investigate multiple variants of two phases on AMT dataset, shown in Fig.~\ref{fig:exp:crowd:ab}.
We explore $3$ approaches to incorporate source-specific representation in the decoupling phase (DP). \texttt{CRF} means the traditional approach as Eq.~\ref{eq:crf} while \texttt{DP(1+2)} is for our method as Eq.~\ref{eq:df:crf}. \texttt{DP(1)} only applies source representations $\bA^{(k)}$ to the emission score $\bU$ while \texttt{DP(2)} only transfers the transition matrix $\bM$.
We can observe from the result that both variants can improve the result.
The underlying model keeps more consensus knowledge if we extract annotator-specific bias on sentence encoding and label transition.
We also compare $4$ methods of generating supervision targets in the aggregation phase (AP). 
\texttt{OMV} uses majority voting of original annotations, while \texttt{PMV} substitutes them with model prediction learned from DP. \texttt{AMV} extends the model by using all prediction, while \texttt{AWV} uses majority voting weighted by each annotator's training $F1$ score.
The results show the effectiveness of \texttt{AWV}, which could augment training data and well approximate the ground truth to supervise the attention module for estimating the expertise of annotator on the current sentence.
We can also infer labels on the test set by conducting \texttt{AWV} on predictions of the underlying model with each annotator-specific components.
However, it leads to heavy computation-consuming and unsatisfying performance, whose test $F1$ score is $77.35(\pm 0.08)$.
We can also train a traditional BLSTM-CRF model with the same \texttt{AMV} labels. Its result is $78.93(\pm 0.13)$, which is lower than \textsc{ConNet} and show the importance of extracted source-specific components.



\noindent\textbf{Performance on simulated datasets.} To analyze the impact of annotator quality, we split the origin train set into $z$ folds and each fold could be used to train a CRF model whose reliability could be represented as $r=1/z$ assuming a model with less training data would have stronger bias and less generalization. We tried $5$ settings where $z=\{5,10,15,30,50\}$ and randomly select $5$ folds for each setting. When the reliability level is too low, \ie~$1/50$, only the base model is used for prediction without annotator representations.
Shown in Fig.~\ref{fig:exp:crowd:simu}(a), \textsc{ConNet} achieves significant improvements over \texttt{MVT-SLM} and competitive performance as \texttt{Crowd-Cat}, especially when annotators are less reliable.

Regarding the annotator quantity, we split the train set into $50$ subsets ($r=1/50$) and randomly select $\{5, 10, 15, 30, 50\}$ folds for simulation.
Fig.~\ref{fig:exp:crowd:simu}(b) shows \textsc{ConNet} is superior to baselines and able to well deal with many annotators while there is no obvious relationship between the performance and annotator quantity in baselines. 
We can see the performance of our model increases as the number of annotators and, regardless of the number of annotators, our method consistently outperforms than other baselines.

\subsection{Cross-Domain Adaptation Performance}
\label{sec:exp:domain}

The results of each task on each domain are shown in Tab.~\ref{tab:app:domain}. We can see that \textsc{ConNet} performs the best on most of the domains and achieves the highest average score for all tasks.
We report the accuracy for POS tagging and classification, and the chunk-level $F1$ score for NER. We can see that \textsc{ConNet} achieves the highest average score on all tasks. \texttt{MTL-MVT} is similar to our decoupling phase and performs much worse. Naively doing unweighted voting does not work well. 

The attention can be viewed as implicitly doing weighted voting on the feature level. 
\texttt{MTL-BEA} relies on a probabilistic model to conduct weighted voting over predictions, but unlike our approach, its voting process is independent from the input context. It is probably why our model achieves higher scores. This demonstrates the importance of assigning weights to domains based on the input sentence. 

\texttt{Tri-Training} trains on the concatenated data from all sources also performs worse than \textsc{ConNet}, which suggests the importance of a multi-task structure to model the difference among domains. 
The performance of \texttt{Crowd-Add} is unstable (high standard deviation) and very low on the NER task, because the size of the crowd vectors is not controllable and thus may be too large. On the other hand, the size of the crowd vectors in \texttt{Crowd-Cat} can be controlled and tuned. 
These two methods leverage domain-invariant knowledge only but not domain-specific knowledge and thus does not have comparable performance.

\begin{figure}[t]
\centering
\includegraphics[width=0.65\linewidth]{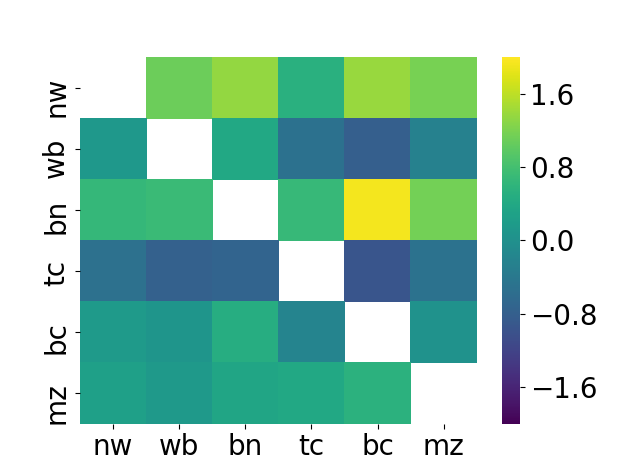}
\caption{\small \textbf{Heatmap of averaged attention scores} from each source domain to each target domain. \vspace{-10pt}}
\label{fig:att_heatmap}
\end{figure}
\subsection{Analyzing Learned Attention }
\label{sec:app:att}

We analyzed the attention scores generated by the attention module on the OntoNotes dataset. For each sentence in the target domain we collected the attention score of each source domain, and finally the attention scores are averaged for each source-target pair. Fig.~\ref{fig:att_heatmap} shows all the source-to-target average attention scores. 
We can see that some domains can contribute to other related domains. For example, bn (broadcast news) and nw (newswire) are both about news and they contribute to each other; bn and bc (broadcast conversation) are both broadcast and bn contributes to bc; bn and nw both contributes to mz (magzine) probably because they are all about news; wb (web) and tc (telephone conversation) almost make no positive contribution to any other, which is reasonable because they are informal texts compared to others and they are not necessarily related to the other. Overall the attention scores can make some sense. It suggests that the attention is aware of relations between different domains and can contribute to the model.


\section{Conclusion}

In this paper, we present \textsc{ConNet}  for learning a sequence tagger from multi-source supervision. 
It could be applied in two practical scenarios: learning with crowd annotations and cross-domain adaptation. 
In contrast to prior works, \textsc{ConNet} learns fine-grained representations of each source which are further dynamically aggregated for every unseen sentence in the target data.
Experiments show that our model is superior to previous crowd-sourcing and unsupervised domain adaptation sequence labeling models. 
The proposed learning framework also shows promising results on other NLP tasks like text classification.

\section*{Acknowledgements}
This research is based upon work supported in part by the Office of the Director of National Intelligence (ODNI), Intelligence Advanced Research Projects Activity (IARPA), via Contract No. 2019-19051600007, NSF SMA 18-29268, and Snap research gift. The views and conclusions contained herein are those of the authors and should not be interpreted as necessarily representing the official policies, either expressed or implied, of ODNI, IARPA, or the U.S. Government. We would like to thank all the collaborators in USC INK research lab for their constructive feedback on the work.

\bibliography{acl2019}
\bibliographystyle{acl_natbib}

\appendix
\clearpage
\section{Analysis of ConNet with BLSTM Encoder} 

\subsection{Case study on learning with crowd annotations}
\label{sec:app:case}

\begin{figure*}[htbp]
    \centering
    \includegraphics[width=\linewidth]{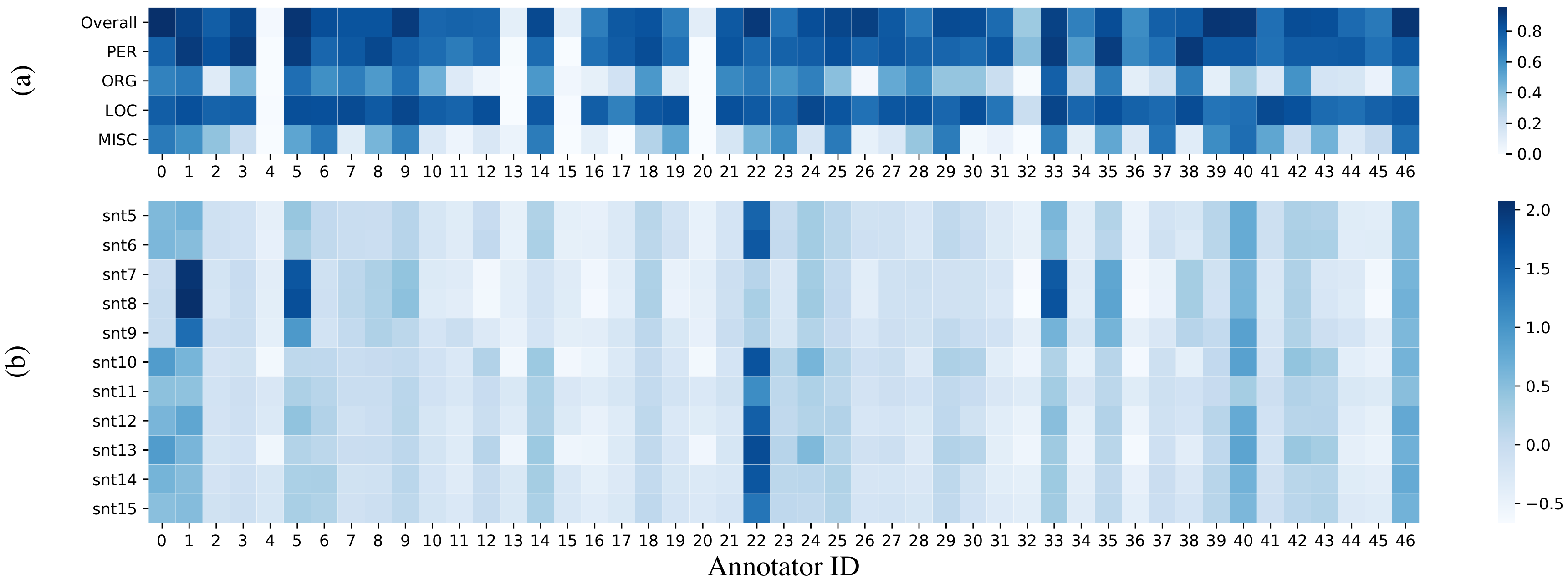}
    \caption{Visualizations of (a) the expertise of annotators; (b) attention weights for additional sample sentences to Fig.~\ref{fig:exp:crowd:caseheat}. Details of samples are described in Tab.~\ref{tab:app:cases}.
    }
    \label{fig:app:caseheat}
\end{figure*}

\begin{table}[htbp]
    \centering
    \begin{tabular}{|c|m{0.9\columnwidth}|}
    \hline
        1 & Defender \PER{Hassan Abbas} rose to intercept a long ball into the area in the 84th minute but only managed to divert it into the top corner of \PER{Bitar} 's goal . \\
        \hline
        2 & \ORG{Plymouth} 4 \ORG{Exeter} 1 \\
        \hline
        3 & Hosts \LOC{UAE} play \LOC{Kuwait} and \LOC{South Korea} take on \LOC{Indonesia} on Saturday in Group A matches . \\
        \hline
        4 & The former \MISC{Soviet} republic was playing in an \MISC{Asian Cup} finals tie for the first time . \\
        \hline
        5 & \PER{Bitar} pulled off fine saves whenever they did . \\ \hline
        6 & \PER{Coste} said he had approached the player two months ago about a comeback . \\ \hline
        7 & \ORG{Goias} 1 \ORG{Gremio} 3 \\ \hline
        8 & \ORG{Portuguesa} 1 \ORG{Atletico Mineiro} 0 \\ \hline
        9 & \LOC{Melbourne} 1996-12-06 \\ \hline
        10 & On Friday for their friendly against \LOC{Scotland} at \LOC{Murrayfield} more than a year after the 30-year-old wing announced he was retiring following differences over selection . \\ \hline
        11 & Scoreboard in the \MISC{World Series} \\ \hline
        12 & Cricket - \MISC{Sheffield Shield} score . \\ \hline
        13 & `` He ended the \MISC{World Cup} on the wrong note , " \PER{Coste} said . \\ \hline
        14 & Soccer - \ORG{Leeds} ' \PER{Bowyer} fined for part in fast-food fracas . \\ \hline
        15 & \ORG{Rugby Union} - \PER{Cuttitta} back for \LOC{Italy} after a year . \\ \hline
        16 & \LOC{Australia} gave \PER{Brian Lara} another reason to be miserable when they beat \LOC{West Indies} by five wickets in the opening \MISC{World Series} limited overs match on Friday . \\ 
        \hline
    \end{tabular}
    \caption{Sample instances in Fig.~\ref{fig:exp:crowd:caseheat} and Fig.~\ref{fig:app:caseheat} with NER annotations including \texttt{PER} (red), \texttt{ORG} (blue), \texttt{LOC} (violet) and \texttt{MISC} (orange).}
    \label{tab:app:cases}
\end{table}

To better understand the effect and benefit of \textsc{ConNet}, we do some case study on AMTC real-world dataset with $47$ annotators. 
We look into some more instances to investigate the ability of attention module to find right annotators in Fig.~\ref{fig:app:caseheat} and Tab.~\ref{tab:app:cases}. Sentence 1-12 contains a specific entity type respectively while 13-16 contains multiple different entities. Compared with expertise of annotators, we can see that the attention module would give more weight on annotators who have competitive performance and preference on the included entity type. Although top selected annotators for \texttt{ORG} has relatively lower expertise on \texttt{ORG} than \texttt{PER} and \texttt{LOC}, they are actually the top three annotators with highest expertise on \texttt{ORG}.

\begin{table*}[htbp!]
\centering
\scalebox{0.9}{
\begin{tabularx}{1.1\textwidth}{lcccXcX}
    \toprule  
    \multirow{2}{*}{\textbf{Methods}}&
    \multicolumn{3}{c}{ \textbf{AMTC}}& \multicolumn{3}{c}{\textbf{UD}}\\
    \cmidrule(lr){2-4} \cmidrule(lr){5-7}
     & Precision(\%) & Recall(\%) & F1-score(\%) & & Accuracy(\%) &  \\
    \midrule  
    {MVT-SLM}  &  72.21($\pm$1.63)  &  51.72($\pm$3.58)  &  60.21($\pm$1.87) & & 87.23($\pm$0.51) & \\
    {Crowd-Add}~\citep{Nguyen2017ACL}  &  75.32($\pm$1.41)  &  50.80($\pm$0.30)  &  60.68($\pm$0.67) & & 88.20($\pm$0.36) & \\
    \midrule
    \textsc{ConNet} (Ours) & \textbf{76.86}($\pm$0.33) & \textbf{56.43}($\pm$3.32) & \textbf{65.05}($\pm$2.32)  & & \textbf{89.27}($\pm$0.31) & \\
    \midrule
    {Gold} (Upper Bound) &  81.24($\pm$1.25)  &  80.52($\pm$0.37)  &  80.87($\pm$0.79) & & 90.45($\pm$0.71) & \\
    \bottomrule 
\end{tabularx}
}
\caption{\small{Performance of methods with Transformer-CRF as the base model on crowd-annotation NER dataset AMTC and cross-domain POS dataset UD.}
}
\label{tab:res:tf}
\end{table*}

\section{Result of ConNet with Transformer Encoder}
\label{sec:app:tf}

To demonstrate the generalization of our framework, we re-implement \textsc{ConNet} and some baselines (\texttt{MTV-SLM}, \texttt{Crowd-Add}, \texttt{Gold}) with Transformer-CRF as the base model. Specifically, the base model takes Transformer as the encoder for CRF, which has shown its effectiveness in many NLP tasks~\citep{vaswani2017attention,devlin-etal-2019-bert}. Transformer models sequences with self-attention and eliminates all recurrence. Following the experimental settings from~\cite{vaswani2017attention}, we set the number of heads for multi-head attention as $8$, the dimension of keys and values as $64$, and the hidden size of the feed-forward layers as $1024$. We conduct experiments with crowd-annotation dataset AMTC on NER task and cross-domain dataset UD on POS task, which are described in Section \ref{sec:exp:data}. Results are shown in Table~\ref{tab:res:tf}. We can see our model outperforms over other baselines in both tasks and applications.

\end{document}